\documentclass[final,3p,times,authoryear]{elsarticle}

\usepackage{amssymb}
\usepackage{amsmath}

\usepackage[autostyle=true]{csquotes}
\usepackage{tcolorbox}
\tcbuselibrary{skins,theorems,breakable}
\usepackage{multirow}
\usepackage{booktabs}
\usepackage{xcolor}
\definecolor{customBlue}{HTML}{1F77B4}
\definecolor{customRed}{HTML}{D62728}
\usepackage{array}
\usepackage{arydshln}

\journal{Arxiv}

\begin{document}
\begin{frontmatter}

\title{Corpus Prevalence of Multiple-Choice Question Options} %

\author[inst1,inst2]{Leonidas Zotos\corref{cor1}}
\cortext[cor1]{Corresponding author.}
\ead{l.zotos@rug.nl}

\author[inst2]{Hedderik van Rijn}

\author[inst1]{Malvina Nissim}

\affiliation[inst1]{organization={Center for Language and Cognition, University of Groningen},
            addressline={Oude Kijk in 't Jatstraat 26}, 
            city={Groningen},
            postcode={9712 EK}, 
            country={The Netherlands}}

\affiliation[inst2]{organization={Department of Experimental Psychology, University of Groningen},
            addressline={Grote Kruisstraat 2/1}, 
            city={Groningen},
            postcode={9712 TS}, 
            country={The Netherlands}}

\begin{abstract}
In recent years, corpus-driven AI methods, such as Large Language Models (LLMs), have seen widespread use in education. While on the surface their abilities look promising for tasks ranging from generating assessment materials to simulating student performance, we should be aware of the subtle nuances of their frequentist nature that might be affecting their behaviour. In this work, we focus on the aspect of corpus frequency in the context of creating high-quality Multiple Choice Questions (MCQs), specifically asking: What if corpus prevalence were enough to identify the correct answer to an MCQ? We propose a computational method of assessing corpus prevalence of MCQ options in large text corpora leveraging textual embeddings using both expert- and machine-generated MCQ sets. The key finding, across three large question sets, is that correct answers, \textit{independently} of the question stem, are significantly more available than incorrect options. Specifically, using Wikipedia as the retrieval corpus, we find that always selecting the most prevalent option leads to scores up to 9.0\% above the random-guess baseline. We also find that MCQ distractors generated by LLMs often show similar patterns of prevalence compared to expert-created options, despite the LLMs' frequentist nature and their training on large collections of textual data. Moreover, we find that corpus prevalence does not necessarily correlate with how recognisable terms are to humans. This highlights the need to better understand how corpora are used in AI-driven methods for education, whether applied directly or indirectly via LLMs.
\end{abstract}


\end{frontmatter}

\section{Introduction}
\label{sec:introduction}
During test-taking, examinees often use \enquote{test-wiseness strategies} of varying complexity to achieve a higher score~\citep{millmanAnalysisTestWiseness1965}. For example, in the context of Multiple-Choice Questions (MCQs), examinees might quickly eliminate absurd options, or use an \enquote{umbrella term} strategy, by which they choose the option that encompasses all known correct options~\citep{mckenna2019multiple,townsStudentUseTestwiseness1993}. Unfortunately, creating MCQs robust to test-wiseness strategies is challenging, as the question designer needs to concurrently pay attention to various aspects of the item. To this end, $19$ \enquote{Item Writing Flaws} have been proposed to function as guidelines for MCQ creators~\citep{Tarrant2006}, including criteria such as the length of the distractors/incorrect choices, the presence of negation in the stem or the plausibility of the distractors. In this work, and adjacent to the \enquote{distractor plausibility} criterion, we explore the relative concept prevalence of MCQ options. For example, considering an MCQ with the options \enquote{Paris}, \enquote{Tallinn}, and \enquote{Antananarivo}, if \enquote{Paris} (which is significantly more prevalent as a concept) is the correct answer, it introduces an undesirable cue for the examinee. 

\begin{figure}[h!]
    \centering
    \includegraphics[width=0.9\linewidth]{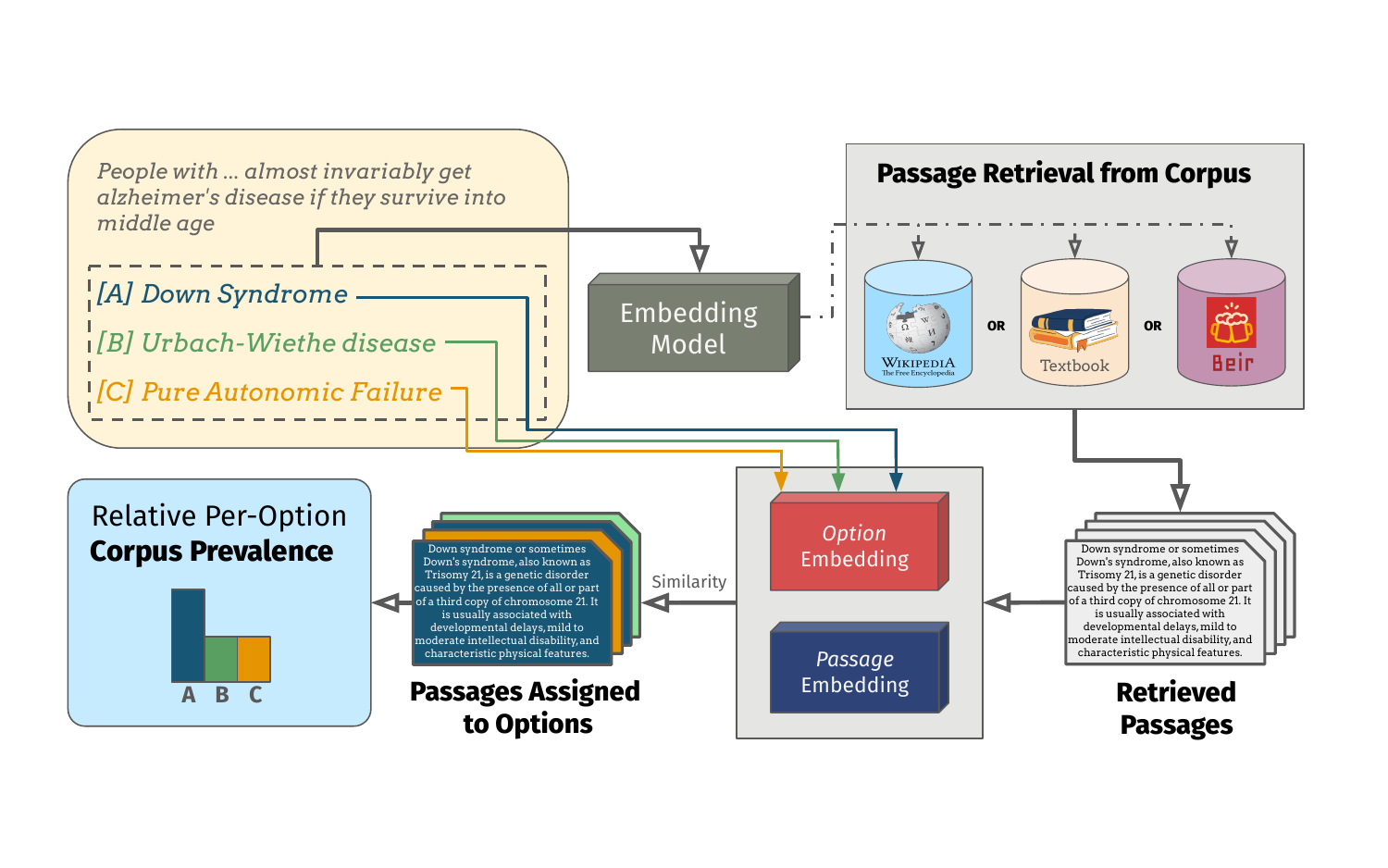}
    \caption{Overview of our approach of evaluating the relative out-of-context prevalence of an MCQ option. First, we compute the textual embedding of the combined MCQ options. Then, a pre-determined number of relevant passages are retrieved from one of three text corpora. Afterwards, the textual embedding of each option is computed separately. The relative per-option prevalence is finally determined by the proportion of retrieved passages that are most similar to each option.}
    \label{fig:choice_corpus_prevalence}
\end{figure}

We further make a distinction between in- and out-of-context relative concept prevalence: While the prevalence of \enquote{Paris} is high out-of-context, relative to the other options, the prevalence of \enquote{Antananarivo} increases when conditioned on the question \enquote{Which of the following is an African capital?}. In other words, we can consider the relative prevalence of MCQ options with or without the question stem. In this work, we propose a prevalence quantification methodology based on the presence of the content of each option in large text corpora (Figure~\ref{fig:choice_corpus_prevalence}). With this approach, we evaluate whether correct MCQ answers show higher relative prevalence compared to incorrect options (also known as distractors), therefore offering a useful test-wiseness strategy to test-takers. We are particularly interested in out-of-context prevalence, as it presents a stronger case as an MCQ quality indication measure (see Section~\ref{sec:in_context_prevalence} for details). Henceforth, we refer to out-of-context prevalence simply as corpus prevalence.

In recent times, various techniques leveraging Large Language Models (LLMs) have been proposed for the generation of MCQ distractors ~\citep{alhazmi-distractor-survey}. This is seemingly a low-hanging fruit, as these tools are widely accessible and the well-defined nature of the task lowers the barrier of entry for educators. However, given the frequentist nature of LLMs and their training regime using large collections of textual data, it is conceivable that distractors generated using them show distinct patterns in terms of corpus prevalence. Therefore, in this work, we further evaluate whether distractors generated by LLMs and humans suffer similarly in terms of relative prevalence.

Last but not least, considering relative prevalence as a potential test-wiseness strategy, we are interested in studying whether this measure is also reflected in human participants. In theoretical terms, this intuition can be linked to the well-established availability heuristic, a strategy by which the likelihood of an event is estimated based on the ease of cognitive retrieval of relevant memories or associations~\citep{availability1973}. In the MCQ context, there can be differences in availability between the options. Although cognitive availability is difficult to measure directly, we evaluate whether correct answers (independent of the question stem) are more recognisable than distractors, and whether this correlates with their corpus prevalence. If so, option recognisability provides a simple test-wiseness strategy that can be estimated via corpus prevalence.

Concretely, our work is structured around three research questions: 

\begin{description}
    \item [\textbf{RQ1}] Is there a difference in relative corpus prevalence between correct answers and distractors in Multiple-Choice Questions?
    \item [\textbf{RQ2}] Do human-created and LLM-generated distractors differ in terms of relative corpus prevalence?  
    \item [\textbf{RQ3}] Is relative corpus prevalence of concepts a proxy for how recognisable they are to the general public?
\end{description}

\section{Related Work}
\label{sec:related_work}
In this section we discuss some literature related to the main topics of the current work. 

\subsection{Generation and Evaluation of MCQ Distractors}
\label{sec:related_work:llms_in_education}
Although LLMs have only existed at their current level of capabilities for a couple years, they have already taken the centre stage in the field of educational technology, offering many opportunities~\citep{vajjala2025opportunities}. In connection to the current work, and focusing on the generation of distractors, most recent methods rely on pre-trained language models that are either fine-tuned on specific data, or, instructed through a prompt. To highlight some examples, \cite{chiang-etal-2022-cdgp} and \cite{wang-etal-2023-distractor} fine-tune BERT-based and text2text~(T5/BART) models respectively to generate distractors for cloze-style MCQs. Similarly, \cite{offerings2020} fine-tune GPT-2 for the task, without specifically focusing on cloze-style questions. All three approaches also apply a distractor-selection process to improve the quality of final distractors.  When using pre-trained LLMs directly, without first fine-tuning them, some work has shown early promise in specific fields using custom instruction prompts \citep{tran_ieee} and by providing in the prompt similar questions with appropriate distractors \citep{bitew2023distractorgenerationmultiplechoicequestions, mcnichols2024automateddistractorfeedbackgeneration}. In the broader context, the extensive use of language models calls for a better understanding of their behaviour for this task, thus motivating \textbf{RQ2}. For a more extensive overview on the use of automated methods for the generation of MCQ distractors, we point the reader to the survey by \cite{alhazmi-distractor-survey}.

Shifting the focus towards MCQ evaluation, and specifically automated techniques to assess distractor quality, \cite{benedetto-distractor-evaluation} broadly divide them into two categories: dynamic, which are based on learners' answers, and static, which are based solely on the textual information of the question item. Of special interest to the current work are the dynamic approaches that use models as proxies for learner behaviour. For example, in the works of \cite{chung-etal-2020-bert} and \cite{offerings2020}, while tackling the task of distractor generation, the authors evaluate their distractors based on the performance of Question-Answering models, assuming that machine behaviour mimics learner behaviour. Moving to static approaches, most recent methods are based on machine-learning techniques that almost exclusively use language models. For example, \citet{qu-etal-2024} train a BERT-based model to predict whether a given distractor is the correct answer to the corresponding question. Following a very different approach, \citet{ghanem2023} use a language model to purposefully generate \enquote{bad distractors}, which are then used to train a model to predict distractor quality. Unfortunately, as \cite{benedetto-distractor-evaluation} points out, there is a general lack in the literature of validating these language-model driven approaches with student data.

\subsection{Concept Familiarity and Frequency}
Most similar to the current work, and particularly relevant to \textbf{RQ3}, is the study by \citet{Kumiko_03431}. There, within the scope of word-meaning acquisition, the authors investigate whether the frequency of words in text corpora correlates with their assessed familiarity. The authors find that words that are more common in corpora are also rated as more familiar, but interestingly, familiar words do not necessarily have high corpus frequency. This latter conclusion is in line with the current study, where we find that corpus prevalence of terms does not consistently correlate with their recognisability. At the same time, \citet{shaoul2013subjective} find that human raters are generally capable of assessing the frequency of words. To offer a potential explanation for this discrepancy, we can look to Grice's maxim of quantity, which posits that utterances provide only as much information as necessary \citep{grice1975logic}. Consistent with this principle, \citet{lin-etal-2012-syntactic} find that universally recognised concepts, such as \enquote{yellow banana}, are explicitly mentioned less frequently in corpora than more specialised or novel terms, such as \enquote{green banana}.

While the aforementioned studies have focused on word frequency, recent developments in information retrieval allow us to capture concept prevalence. Notably, in the landmark work by \citet{karpukhin2020densepassageretrievalopendomain}, the authors show that using text embeddings from a pre-trained BERT language model to retrieve passages relevant to a query can lead to better retrieval performance compared to keyword-based approaches. This is an important shift, as textual embeddings can capture conceptual nuances, such as the fact that \enquote{metro} and \enquote{subway} refer to the same concept. Moreover, because they can capture the semantic meaning of a sentence regardless of its exact phrasing, this approach allows us, for example, to retrieve passages relevant to an entire MCQ.

\subsection{The Frequentist Nature of LLMs}
Although LLMs are increasingly being anthropomorphised \citep{shardlow-etal-2025-exploring}, their output often reveals their frequentist and non-human nature. \citet{hivemind} study the output of numerous LLMs to open-ended queries, finding a high level of homogeneity not only within the same LLM, but also across families of LLMs. In a sense, this is unsurprising, as there is a high overlap in the corpora on which LLMs are trained. At the same time, it also highlights that LLMs, regardless of how creative they can appear, are still statistical in nature. In the context of MCQs, \citet{balepur-etal-2024-artifacts} find that LLMs generally achieve above random-chance accuracy when only shown the options. While the authors explore a few potential ways they manage this (e.g., memorizing questions from the training data or inducing the question), it remains unclear exactly how they achieve such high performance. By extension, the authors also raise questions about the way we assess the true performance of LLMs. This worry is shared by other researchers as well, who have found that LLMs are very sensitive to seemingly arbitrary elements of MCQs, such as the order of the options \citep{wei2024unveiling, pezeshkpour-hruschka-2024-large} or the presence of specific characters, such as a space at the end of the question \citep{mind_the_gap}. In connection with the current work, we consider it important to study how the frequentist nature of LLMs might be affecting their ability to generate distractors, as well as whether the random-guess baseline is actually appropriate to measure their capabilities.

\section{Methodology}
To tackle the outlined research questions, it is necessary to have an appropriate method of determining the relative corpus prevalence of the MCQ options (\textbf{RQ1}). To this end, we use a multi-step retrieval-based approach, described in Section~\ref{sec:determining_option_corpus_prevalence}. Furthermore, the human-created question sets and the altered versions containing LLM-generated distractors are detailed in Sections~\ref{sec:distractors_generation_method} and ~\ref{sec:manually_created_question_sets}, respectively (\textbf{RQ2}). Lastly, Section ~\ref{sec:behavioural_experiment} details the experimental setup used to assess whether relative corpus prevalence proxies how recognisable concepts are to humans (\textbf{RQ3}).

\subsection{Determining Option Corpus Prevalence}
\label{sec:determining_option_corpus_prevalence}
To measure the relative corpus prevalence of the concepts present in the MCQ options we rely on their semantic similarity to passages in large text corpora. Specifically, for a given MCQ, we retrieve a pre-determined number of relevant text passages from a corpus and record how they are semantically distributed across the options of the MCQ. As a hypothetical example, for the MCQ options \enquote{Down syndrome}, \enquote{Urbach-Wiethe's disease} and \enquote{Pure autonomic failure}, we determine each concept's relative prevalence by the proportion of retrieved passages most similar to each concept.

\subsubsection{Knowledge Corpora}
As sources for the passage retrieval, we use three types of corpora: English Wikipedia~\citep{wikipedia}, BEIR, which is a large information retrieval dataset~\citep{thakur2021beir}, and domain-specific open-access textbooks~\citep{geography_textbook, physics_textbook, biology_textbook, chemistry_textbook, immunology, biopsychology_dommett, biopsychology_hove}. 

Using Wikipedia is a natural choice, as it contains large amounts of highly curated factual information. On the other side of the spectrum, the passages in BEIR cover a much broader scope (e.g., news articles, forum threads and financial investment articles), albeit with less detail. Lastly, using textbooks that are much smaller but domain-specific is a logical choice, as they better represent the knowledge that a student is normally exposed to. As detailed in Section~\ref{sec:manually_created_question_sets}, this work focuses on the domains of Biopsychology, Immunopharmacology, and General High-School Science, using relevant textbooks for each. Table \ref{tab:corpora_overview} details the corpora used in this work.

\begin{table}[h!]
\caption{Text corpora used for passage retrieval. When using textbooks as our retrieval corpora, we focus on the combination of textbooks relevant to the respective domain. All corpora are publicly accessible.}
\centering
\begin{tabular}{c >{\centering\arraybackslash}m{4.15cm} m{7cm} c}
\toprule
\textbf{Corpus} & \textbf{Domain} & \textbf{Source} & \textbf{\# Of Passages} \\
\midrule
\begin{tabular}[c]{@{}c@{}}\textbf{English}\\\textbf{Wikipedia}\end{tabular} & General & Wikipedia: The Free Encyclopedia \citep{wikipedia} & 41.5M \\
\midrule
\textbf{BEIR} & Very General & BEIR: A Heterogeneous Benchmark for Zero-shot Evaluation of Information Retrieval Models \citep{thakur2021beir} & 50.5M \\
\midrule
\multirow{7}{*}{\textbf{Textbooks}} & University Biopsychology & Introduction to biological psychology \citep{biopsychology_dommett} \newline Biological psychology \citep{biopsychology_hove} & 1446 \\
\cdashline{2-4}
 & University Immunopharmacology & Immunology \citep{immunology} & 542 \\
\cdashline{2-4}
 & High School Science & College Physics \citep{physics_textbook} \newline Concepts of Biology \citep{biology_textbook} \newline Introduction to Chemistry: General, Organic, and Biological \citep{chemistry_textbook} \newline Introduction to Human Geography \citep{geography_textbook} & 3828 \\
\bottomrule
\end{tabular}
\label{tab:corpora_overview}
\end{table}

\subsubsection{Retrieving and Assigning Passages}
In order to retrieve passages relevant to a given MCQ, we leverage semantic similarity between the MCQ options and the passages in the corpus. For this, relying on textual embeddings instead of n-gram frequency is preferable, as the former are more flexible and better capture conceptual nuance, for example capturing the equivalence between \enquote{flu} and \enquote{influenza}. Specifically, we use Cohere's \enquote{Embed v3}~\citep{cohereEmbedv3} model to compute textual embeddings for each passage in the corpus and the combined options of each MCQ (e.g., \enquote{Paris Tallinn Antananarivo}). We only use the combined options as the retrieval query, excluding the question stem, as to measure out-of-context, instead of in-context prevalence. This choice also reduces the bias towards the correct answer: building on the earlier example, the presence of \enquote{France} in the query will lead to the retrieval of more passages related to Paris, thus obstructing its genuine corpus prevalence. It is worth noting that the presence of all options during retrieval still offer some context, albeit less than the presence of the question stem. We explore the effect of the retrieval query in more detail in \ref{sec:in_context_prevalence}. Lastly, in \ref{sec:number_of_retrieved_passages} we also explore the effect of the number of passages retrieved, by retrieving between $1$ and $25$ passages for each MCQ. This allows us to evaluate the robustness of our methodology against the imbalance of certain domains in corpora (e.g., general geography content is significantly more prevalent than specialised immunopharmacology). An alternative approach to retrieving a pre-determined number of passages is to use a semantic similarity threshold. However, we avoid this approach because an optimal threshold is typically domain- and question-specific, making it difficult to set dynamically.

Once the passages relevant to the MCQ options are retrieved, each is assigned to one of the options based on semantic similarity. To achieve this, we compute the textual embedding of the passage and each option. The passage is then assigned to the option with the highest cosine similarity, leading to a distribution of passages across the options (Figure~\ref{fig:choice_corpus_prevalence}).

\subsection{Question Sets}
\label{sec:manually_created_question_sets}
We conduct our main experiments using three exam question sets covering the topics of Biopsychology, Immunopharmacology and high-school level Science. All question sets target factual knowledge and are in English. We focus on items targeting factual knowledge, where corpus prevalence differences between options are most likely to be pronounced, in contrast to reasoning or mathematics-oriented MCQs.

Both the Biopsychology and Immunopharmacology question sets originate from undergraduate courses taught at our institution. The Biopsychology set includes 319 three-option and 278 four-option MCQs from the Psychology programme, covering the textbook \enquote{Biological Psychology} \citep{Kalat_2016}. We further divide this question set into two subsets based on the number of options, hereafter referred to as \enquote{Biopsychology-3} and \enquote{Biopsychology-4}. We make this distinction because the sets were created by different authors and corpus prevalence is measured per-option.

In comparison, the Immunopharmacology dataset contains 489 four-option MCQs from the Pharmacy programme based on \enquote{Basic Immunology} \citep{Abbas2023}. For both question sets, the difficulty of each item is known, quantified through the proportion of students answering each MCQ correctly, using on average 300 and 100 responses for Biopsychology and Immunopharmacology respectively. The relationship between the corpus prevalence of the MCQ choices and question difficulty is explored in Section~\ref{sec:corpus_prevalene_question_difficulty_corr}. In the context of the current study, these question sets are ideal, as they accurately reflect standard factual MCQs used in real-world examinations, albeit in relatively specialised fields. Furthermore, the two sets are private, ensuring that the LLMs used to generate alternative distractors (as described in Section~\ref{sec:distractors_generation_method}) have not encountered the human-created distractors during their training. 

Moreover, we also use SciQ~\citep{sciq2017}, a dataset containing a broad range of 4-option science questions. While the full public question set contains approximately 13 thousand questions, we focus on the \enquote{test} split which consists of 1000 MCQs. Only using questions in the \enquote{test} split is crucial, as they carry a lower risk of being present in training data of the LLMs we used to generate alternative distractors. It is worth noting that as the original work mentions, in the creation of this dataset, crowd-workers formulated their own distractors, but they also had the option to use suggested distractors generated using a small Random Forest model. This is an important characteristic, as crowd-workers were generally not domain experts capable of creating pedagogically valid distractors themselves. Lastly, student performance data is not available for the SciQ dataset.

As mentioned earlier, we limit the scope of this study to MCQs targeting factual knowledge. With this scope in mind, we ensured that all MCQ options contained at least one term representing a concept. This selection criterion led to the removal of 61 MCQs from the SciQ dataset where the distractors were numerical (e.g., \enquote{one}, \enquote{two}, \enquote{three}) or contained no clear concepts (e.g., \enquote{left}, \enquote{right}, \enquote{center}). It is worth noting that MCQs with distractors that were full sentences (e.g., \enquote{Via the efferent lymphatic vessels.}) were not excluded, as long as at least one term was present. Some example items from all three datasets are presented in Table~\ref{tab:dataset_samples}.

\begin{table}[h!]
\footnotesize
\caption{Sample questions from the question sets. Correct answers are highlighted in \textcolor{customBlue}{blue}.} 
\centering
\setlength{\tabcolsep}{5pt}
\begin{tabular}{l p{8cm}}
\toprule
\textbf{Dataset} & \textbf{Question and Options} \\
\midrule
\multirow{5}{*}{\textbf{Biopsychology}} 
 & It is easiest to read fine newspaper print when staring directly at it because the fovea ...\\
 & A) surrounds the optic nerve \\
 & \textcolor{customBlue}{B) has many photoreceptors packed there} \\
 & C) is closest to the pupil \\
\midrule
\multirow{6}{*}{\textbf{Immunopharmacology}} 
 & Which cells are part of the adaptive immune system? \\
 & \textcolor{customBlue}{A) Plasma cells} \\
 & B) Monocytes \\
 & C) Macrophages \\
 & D) Mast cells \\
\midrule
\multirow{6}{*}{\textbf{SciQ}}
 & Which type of tree is dominant in temperate forests? \\
 & A) fungus \\
 & B) vines \\
 & \textcolor{customBlue}{C) deciduous} \\
 & D) shrubs \\
\bottomrule
\end{tabular}
\label{tab:dataset_samples}
\end{table}

\subsection{Generating Alternative Distractors using LLMs}
\label{sec:distractors_generation_method}
We further aim to evaluate whether human- and LLM-generated options show similar patterns in terms of corpus prevalence. To this end, we employ three LLMs from the Qwen~3 family~\citep{qwen3}, specifically the instruction-tuned/chat variants with $8$, $30$, and $80$ billion parameters. Using three LLMs of the same family but differing in size allows us to also explore whether any observed corpus prevalence effects are mitigated or enhanced by using more powerful LLMs. We chose to focus on the Qwen~3 family as its models are capable, offer the aforementioned size scaling, and are more lightweight compared to proprietary closed-source alternatives. Furthermore, to evaluate whether our findings generalise to LLMs of other families, we extend our experiments to the Gemma~4-31B~\citep{gemma4} and GPT-OSS-120B~\citep{gpt-oss} models. 

To generate alternative distractors for a given MCQ, a simple generation prompt is used that includes the question stem, the correct answer, and the expected number of distractors (Figure~\ref{fig:distractor_generation_prompt}). In approximately 0.5\% of cases, the wrong number of distractors was generated, the correct answer was presented as a distractor, or generated distractors were repeated. When this occurred, generation was repeated until the conditions were satisfied. While curated prompts and in-context learning can yield pedagogically superior distractors~\citep{alhazmi-distractor-survey}, we rely on a simpler yet carefully crafted instruction that reflects natural model interaction.

\begin{figure}[h!]
    \footnotesize
    \centering
        \tcbset{colback=white,colframe=black,
        fonttitle=\bfseries}
        \begin{tcolorbox}[enhanced,title=Instruction Prompt for Generation of Distractors,
            frame style={color=gray}
        ]        
        Provide \texttt{n} good distractors for this Multiple-Choice Question. The correct answer is \enquote{\texttt{answer}}. Provide them in a \enquote{\texttt{boxed}} environment, separated by a vertical bar. Do not include explanations. Do not include the correct answer as a distractor and make sure to generate \enquote{\texttt{n}} distractors.
        
        Question: \enquote{\texttt{Question Stem}}
        
        Alternative Distractors: 
        \end{tcolorbox}
    \caption{LLM instruction used to generate \texttt{n} alternative distractors. The model is instructed to output the distractors in a \enquote{\texttt{boxed}} environment, separated by vertical bars to facilitate automatic extraction. We empirically find that repetition in the instruction leads to better instruction-following, especially for the less-capable Qwen3-8b.}
    \label{fig:distractor_generation_prompt}
\end{figure}

\begin{figure}[h!]
    \centering
    \includegraphics[width=0.7\linewidth]{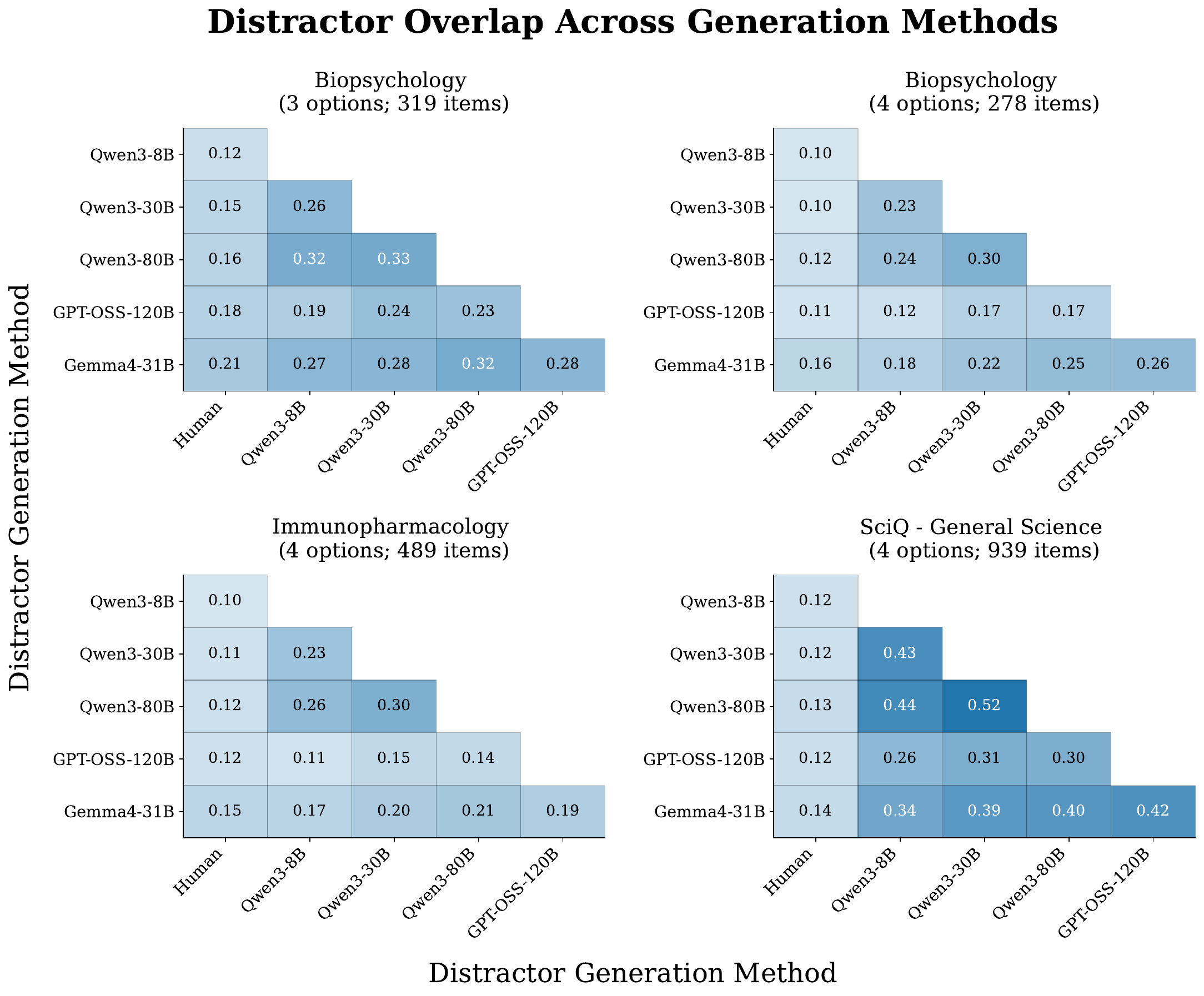}
    \caption{Average proportion of distractors that are identical between the human and LLM-created distractors, across the four question sets.}
    \label{fig:distractor_overlap}
\end{figure}

Given the constrained nature of the task and the previously found homogeneity in LLM output~\citep{hivemind}, it is important to examine the extent to which the generated distractors overlap with the human-generated distractors, as well as with each other. As shown in Figure~\ref{fig:distractor_overlap}, all LLMs evaluated in this study generate distractors that are similar to one another but generally dissimilar to the human-generated distractors. This dissimilarity also holds for the public SciQ question set, suggesting that the LLMs were not exposed to the human-generated distractors during training. Moreover, as expected, the Qwen models—which were likely trained on the same data—generate more similar distractors compared to the other LLMs. Lastly, we observe that LLM-generated distractors show the highest overlap for the SciQ question set, suggesting that these questions have fewer reasonable distractor options.

\section{Behavioural Experiment}
\label{sec:behavioural_experiment}
To address RQ3, we conducted a behavioural experiment to evaluate whether an MCQ option's relative corpus prevalence proxies how easily human participants recognise it. The experiment followed a within-subjects design, varying distractor source (human- or LLM-created distractors) and question set (drawing questions from the four question sets described in Section~\ref{sec:manually_created_question_sets}). We repeated the experiment twice, sampling items from the question sets either using a \enquote{random} or \enquote{most prevalent} sampling method. The latter prioritises items whose correct answers exhibit the highest corpus prevalence (as described in Section~\ref{sec:determining_option_corpus_prevalence}), where we expect any findings regarding recognisability to be accentuated.

\subsection{Materials and Stimuli}
The experiment is conducted online in the form of a questionnaire. From each question set, we sample 50 items using either a \enquote{random} or \enquote{most prevalent} sampling method. We extend the setup for the sets with LLM-generated distractors~\footnote{For this, we chose the Qwen3-80b subset. Because its distractors overlap most with those generated by the other models (as seen in Section~\ref{sec:distractors_generation_method}), it serves as an appropriate representative for the other LLMs.}, resulting in a total of 400 items per sampling method. For each item, the participant is shown just the options without the question stem, accompanied with the instruction \enquote{If we surveyed a large, random group of people, which term would be recognized by the highest number of them?}.  We assess term recognition indirectly, similar how subjective word frequency is measured in literature~\citep{shaoul2013subjective}. To avoid any confusion for the participants, for this experiment we only use MCQs whose options are 1- or 2-word terms instead of sentences. The terms in each trial were presented horizontally, and the participants submitted their answers through standard radio buttons. In order to gauge whether participants are paying attention to the task we create two \enquote{attention check} items for each set that fit the subject but contain a term that is significantly more recognisable (e.g., \enquote{stomach}, \enquote{thymus}, \enquote{periosteum}). Naturally, these attention check items are only used to filter out participants and are omitted from the analysis of the results.

\subsection{Participants}
Participants were recruited through Prolific, an online participant recruitment platform \citep{prolific2026}. As the items are all in English, we only recruited participants with English as their first language, whose current country of residence was the United Kingdom, United States, Australia or New Zealand. Moreover, only participants with a 100\% approval rating on the platform were recruited.

Of the 115 participants who were recruited, nine failed at least four out of the eight attention checks and were thus removed from the study. Of the remaining 106 participants, 56 (32 females, 23 males, 1 other; $M_{\text{age}} = 33.2$, $SD = 11.0$) were shown the items following the random sampling method, and 50 (22 females, 28 males; $M_{\text{age}} = 37.0$, $SD = 11.9$) were shown the items following the most prevalent sampling method. Participants were compensated through Prolific, following its suggested hourly payment rate. 

\subsection{Procedure}
Regardless of the item sampling method, participants followed the same procedure. Before the experiment, they were provided with a digital informed consent explaining all their rights during participation. Once participants gave consent to participate in the experiment, complete instructions for the experiment were given to participants, as shown in Figure~\ref{fig:experiment_instructions}. A practice session followed to familiarise the participants with the task, consisting of four trials that were unrelated to the four question sets (e.g., \enquote{Sedan}, \enquote{Coupe}, \enquote{Limousine}). 

\begin{figure}[h!]
    \footnotesize
    \centering
        \tcbset{colback=white,colframe=black,
        fonttitle=\bfseries}
        \begin{tcolorbox}[enhanced,title=Information given to the participants,
            frame style={color=gray}
        ]        
        In this task, we are interested in how people predict the knowledge of the general public. For each question, you will see a few different terms. Your job is to guess which of those terms is most widely recognised by the general public. For each set, we want you to answer the following question: \enquote{If we surveyed a large, random group of people, which term would be recognized by the highest number of them?}.

        We have deliberately included a very wide variety of topics. Some sets will contain common, everyday words. Other sets will contain highly specialised and obscure terms. \textbf{It is completely expected that you will not know what some of these words mean. This is not a test of your vocabulary!} When you encounter a set of difficult or unfamiliar words, just go with your first instinct and select the one that feels like it might be slightly more recognisable to the average person. We do not expect you to spend more than a couple of seconds per item.

        Study Structure: The experiment will take approximately 20 minutes to complete. It consists of a short practice trial, followed by 4 blocks of 52 items each.
        \\
        Note: We have embedded multiple \textbf{attention checks} throughout the survey to verify response quality and confirm eligibility for payment.

        \end{tcolorbox}
    \caption{Information given to the participants prior to the start of the experiment.}
    \label{fig:experiment_instructions}
\end{figure}

Following the practice trial, the main experiment started. To mitigate fatigue, participants evaluated a randomly sampled subset of 50 items per question set, consisting of both human- and LLM-generated items. The trials were divided into four blocks (one per question set) and separated by brief rest periods. Two attention-check items matching the theme of the block were randomly placed into each block. The presentation order of the blocks, the trials within them, and the terms within each trial were randomised. On average, the experiment lasted approximately 22 minutes ($M = 21.9$, $SD = 5.3$).

\section{Results}
As discussed in Section~\ref{sec:determining_option_corpus_prevalence}, an important variable in our methodological setup is the number of retrieved passages for each question item. For all experiments presented in this section, we retrieve five passages per item, as we find that retrieving more does not change the corpus prevalence measure (while being much more computationally expensive), and retrieving fewer leads to more unstable results. An analysis of the effect of the number of retrieved passages on the measured corpus prevalence is presented in \ref{sec:number_of_retrieved_passages}. 

Furthermore, as discussed in Section~\ref{sec:introduction}, we consider the relative corpus prevalence of the options out-of-context, that is to say, without considering the question stem. Expectedly, measuring corpus prevalence in-context primes the passage retrieval towards the correct answer, leading to a larger difference in the proportion of passages retrieved relevant to the correct answer compared to the distractors. The corresponding results of this section using in-context prevalence are presented in \ref{sec:in_context_prevalence}.

\subsection{Relative Corpus Prevalence}
\label{sec:relative_corpus_prevalence}
\begin{figure}[h!]
    \centering
    \includegraphics[width=1\linewidth]{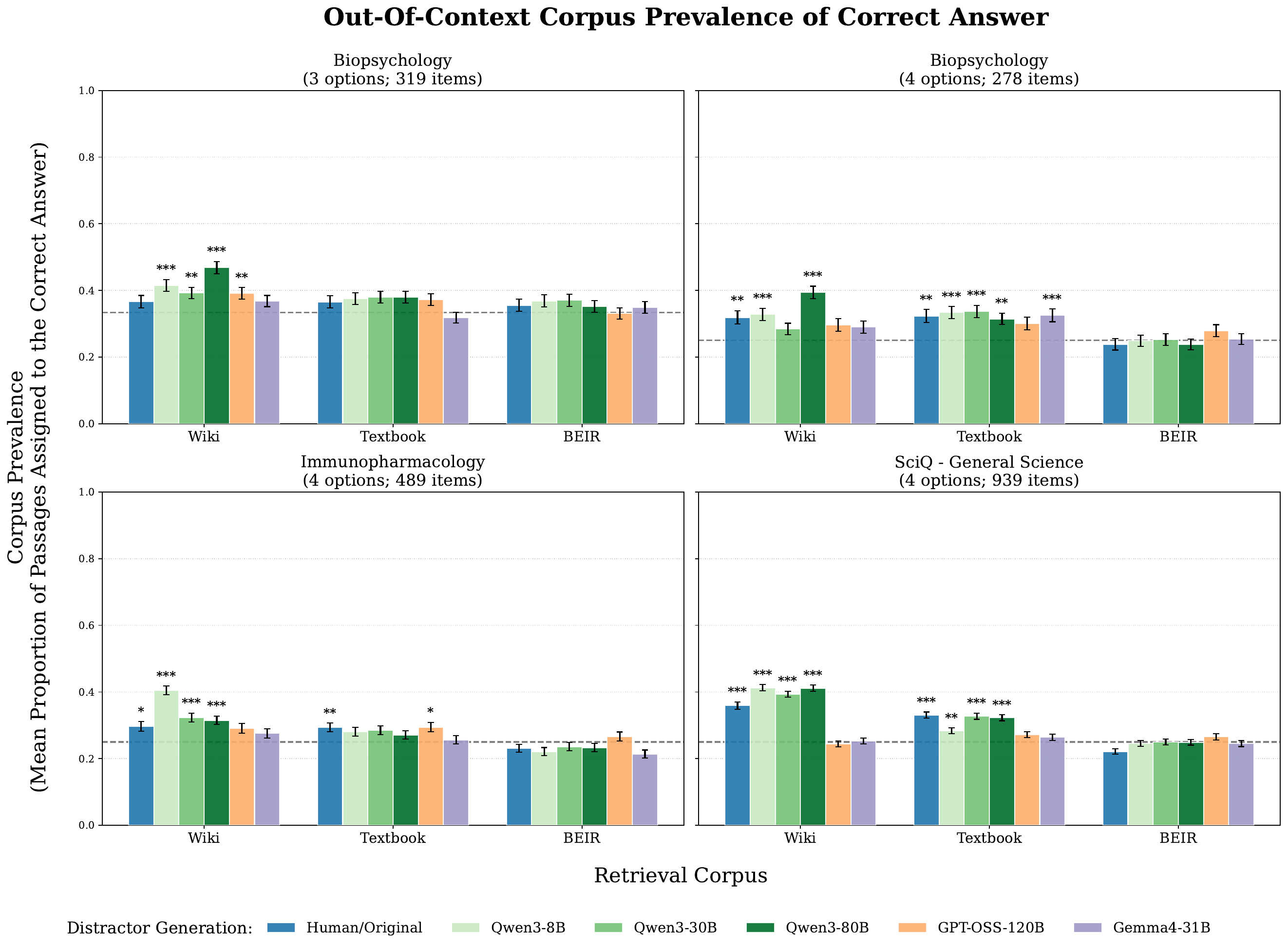}
    \caption{Relative out-of-context corpus prevalence of the correct answer of MCQs varying the distractor generation method and the retrieval corpus. Dashed lines indicate the baseline random-assignment proportion. Asterisks indicate the strength of evidence for the alternative hypothesis based on Bayes Factors: * $BF_{10} > 3$ (moderate), ** $BF_{10} > 10$ (strong), and *** $BF_{10} > 100$ (extreme).}
    \label{fig:corpus_prevalence_out_of_context}
\end{figure}

To address \textbf{RQ1} and \textbf{RQ2} regarding whether the correct answers of MCQs are more prevalent relative to the human- and LLM-generated distractors, we study the average proportion of retrieved passages that are most similar to them. Specifically, for the 3- and 4-option MCQs, we evaluate whether the proportion of passages assigned to the correct answer exceeds the baseline random-assignment proportions of 0.33 and 0.25, respectively. We use a two-sided Bayesian one-sample T-Test to assess whether the corpus prevalence of the correct answer is significantly different than the random-assignment proportion (Figure~\ref{fig:corpus_prevalence_out_of_context}). This analysis is repeated across all three corpora.

Regarding \textbf{RQ1}, we find that for all question sets the correct answer tends to be more prevalent on average than the human-generated distractors in the Wikipedia and Textbook corpora, with the difference being significant for all but the Biopsychology-3 question sets. In contrast, we find that the correct answer is not consistently more prevalent in the BEIR corpus compared to the distractors, which might be due to BEIR having a broader focus lacking specialised passages relevant to the question items. 

Addressing \textbf{RQ2}, we observe from this analysis that while the distractors generated by GPT-OSS-120B and Gemma4-31B are not consistently more or less prevalent than the correct answers in the tested corpora, Qwen-generated distractors exhibit similar corpus prevalence as human-generated distractors. Additionally, we do not find differences across the Qwen models of different sizes, suggesting that increased general capabilities do not lead to the generation of distractors with a corpus prevalence more similar to that of the correct answer.

While this analysis shows that the correct answer tends to be more prevalent in corpora, we are also interested in the proportion of question items where this pattern is present. In other words, with LLM evaluation in mind, we want to determine the baseline accuracy a system would achieve by simply selecting the most prevalent option. As shown in Table~\ref{tab:accuracy_majority_out_of_context}, when using Wikipedia as the retrieval corpus, these gains range from 0.9\% to 9.0\% for the human-generated sets, and are often higher for the sets generated by the Qwen models. In line with the findings presented previously in Figure~\ref{fig:corpus_prevalence_out_of_context}, the proportions for the Gemma4-31b and GPT-120b sets fall close to or below the baseline. As established earlier, this gain is significantly higher when the question stem is included during the corpus prevalence measurement (\ref{sec:in_context_prevalence}). This finding raises concerns about the inherent advantage that the frequentist nature of LLMs provides when answering questions where this prevalence discrepancy is present.

\begin{table}[h!]
\caption{Proportion of items where the correct answer has higher out-of-context prevalence in the Wikipedia than the distractors. The increase (or decrease) over the baseline (33.3\% for 3 options, 25.0\% for 4 options) is denoted in brackets.}
\centering
\setlength{\tabcolsep}{5pt}
\begin{tabular}{ccccc}
\toprule
\multirow{2}{*}{\textbf{Distractors Generator}} & \multicolumn{4}{c}{\textbf{\% Correct Answer as Most Prevalent (Wikipedia)} $\uparrow$} \\
\cmidrule(lr){2-5}
 & \shortstack[c]{\textbf{Biopsychology} \\ (3 options)} & \shortstack[c]{\textbf{Biopsychology} \\ (4 options)} & \shortstack[c]{\textbf{Immunopharmacology} \\ (4 options)} & \shortstack[c]{\textbf{SciQ} \\ (4 options)} \\
\midrule
Human/Original & 34.2\% {\scriptsize (+0.9\%)} & 30.9\% {\scriptsize (+5.9\%)} & 27.4\% {\scriptsize (+2.4\%)} & 34.0\% {\scriptsize (+9.0\%)} \\
Gemma4-31b     & 31.3\% {\scriptsize (-2.0\%)} & 24.8\% {\scriptsize (-0.2\%)} & 24.7\% {\scriptsize (-0.3\%)} & 20.4\% {\scriptsize (-4.6\%)} \\
GPT-120b       & 33.2\% {\scriptsize (-0.1\%)} & 25.9\% {\scriptsize (+0.9\%)} & 26.2\% {\scriptsize (+1.2\%)} & 19.4\% {\scriptsize (-5.6\%)} \\
Qwen3-8b       & 37.3\% {\scriptsize (+4.0\%)} & 30.9\% {\scriptsize (+5.9\%)} & 36.2\% {\scriptsize (+11.2\%)} & 38.7\% {\scriptsize (+13.7\%)} \\
Qwen3-30b      & 35.4\% {\scriptsize (+2.1\%)} & 25.2\% {\scriptsize (+0.2\%)} & 28.8\% {\scriptsize (+3.8\%)} & 38.0\% {\scriptsize (+13.0\%)} \\
Qwen3-80b      & 42.0\% {\scriptsize (+8.7\%)} & 38.5\% {\scriptsize (+13.5\%)} & 24.5\% {\scriptsize (-0.5\%)} & 39.0\% {\scriptsize (+14.0\%)} \\
\bottomrule
\end{tabular}
\label{tab:accuracy_majority_out_of_context}
\end{table}

\subsection{Corpus Prevalence and Recognisability}
\label{sec:corpus_prevalence_and_recognisability}
To address \textbf{RQ3}, we evaluate the relation between MCQ options' relative corpus prevalence and their respective recognisability, following the experimental setup described in Section~\ref{sec:behavioural_experiment}. First, to obtain a general overview of this relation, we evaluate the correlation between recognition rate and corpus prevalence using Kendall's $\tau_b$ rank correlation coefficient for all options, regardless of whether they are the correct answer. We selected this non-parametric test because our corpus prevalence metric is restricted to discrete 0.2 increments (representing the proportion of five retrieved passages), which produces tied ranks in the measurements. Moreover, to determine statistical significance, we use Bonferroni correction to account for multiple comparisons. 

As detailed in Section~\ref{sec:behavioural_experiment}, the results are based on the experiment using 100 items from each question set, equally divided between human- and Qwen3-80b-generated distractors, sampling either randomly from the full set or by prioritising items whose correct answers had the highest prevalence in Wikipedia. As shown in Table~\ref{tab:recognition_prevalence_corr}, we observe a somewhat counter-intuitive negative correlation between corpus prevalence and recognition. This correlation is moderate and statistically significant for the 3-option Biopsychology set, as well as the Immunopharmacology set when using LLM-generated distractors. Furthermore, correlations are generally more pronounced in the presence of LLM-generated distractors and when sampling prioritized high-prevalence items, although most of these effects are not statistically significant.

\begin{table}[h!]
\caption{Kendall's $\tau$ correlation between corpus prevalence and recognition of MCQ options. Statistical significance after Bonferroni correction for multiple comparisons is denoted with an asterisk~($\alpha = 0.006$).}
\centering
\setlength{\tabcolsep}{5pt}
\begin{tabular}{llccccc}
\toprule
\multirow{3}{*}{\textbf{Source Question Set}} & \multirow{3}{*}{\shortstack[l]{\textbf{Distractor} \\ \textbf{Generation}}} & \multirow{3}{*}{\textbf{$N$}} & \multicolumn{4}{c}{\textbf{Sampling Method}} \\
\cmidrule(lr){4-7}
 & & & \multicolumn{2}{c}{\textbf{Random}} & \multicolumn{2}{c}{\textbf{Most Prevalent}} \\
\cmidrule(lr){4-5} \cmidrule(lr){6-7}
 & & & \textbf{$\tau_b$} & \textbf{$p$-value} & \textbf{$\tau_b$} & \textbf{$p$-value} \\
\midrule
\multirow{2}{*}{\shortstack[l]{\textbf{Biopsychology} \\ (3 options)}} 
 & Human & 150 & -0.1756 & $3.69 \times 10^{-3}(*)$ & -0.2028 & $9.45 \times 10^{-4}(*)$ \\
 & LLM   & 150 & -0.2431 & $6.16 \times 10^{-5}(*)$ & -0.2755 & $5.54 \times 10^{-6}(*)$ \\
\midrule
\multirow{2}{*}{\shortstack[l]{\textbf{Biopsychology} \\ (4 options)}} 
 & Human & 200 & -0.0495 & $3.59 \times 10^{-1}$ \phantom{(*)} & -0.0754 & $1.65 \times 10^{-1}$ \phantom{(*)} \\
 & LLM   & 200 & -0.1152 & $3.12 \times 10^{-2}$ \phantom{(*)} & -0.0598 & $2.69 \times 10^{-1}$ \phantom{(*)} \\
\midrule
\multirow{2}{*}{\shortstack[l]{\textbf{Immunopharmacology} \\ (4 options)}} 
 & Human & 200 & -0.0979 & $6.57 \times 10^{-2}$ \phantom{(*)} & -0.0077 & $8.87 \times 10^{-1}$ \phantom{(*)} \\
 & LLM   & 200 & -0.1583 & $3.11 \times 10^{-3}(*)$ & -0.1807 & $7.53 \times 10^{-4}(*)$ \\
\midrule
\multirow{2}{*}{\shortstack[l]{\textbf{SciQ} \\ (4 options)}} 
 & Human & 200 & -0.1172 & $2.87 \times 10^{-2}$ \phantom{(*)} & -0.0041 & $9.46 \times 10^{-1}$ \phantom{(*)} \\
 & LLM   & 200 & -0.0266 & $6.22 \times 10^{-1}$ \phantom{(*)} & -0.1311 & $1.59 \times 10^{-2}$ \phantom{(*)} \\
\bottomrule
\end{tabular}
\label{tab:recognition_prevalence_corr}
\end{table}

\begin{figure}[h!]
    \centering
    \includegraphics[width=1\linewidth]{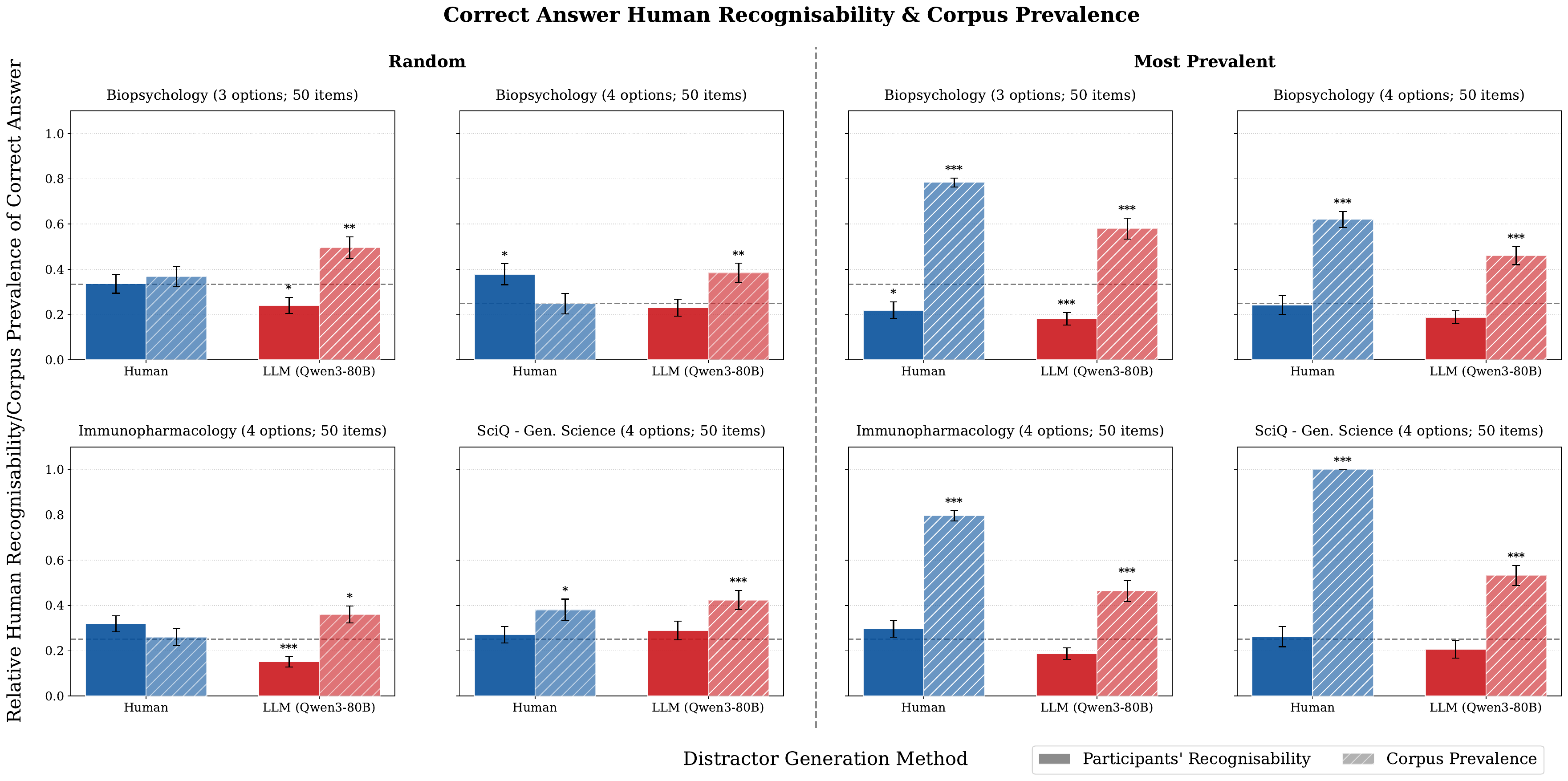}
    \caption{Mean Relative Human Recognisability/Corpus Prevalence of the Correct Answer, using the \enquote{random} and \enquote{most prevalent} sampling methods previously described in Section~\ref{sec:behavioural_experiment}. It is worth noting that the items whose answers have the highest corpus prevalence are determined using the human-created distractors. Asterisks indicate the strength of evidence for the alternative hypothesis based on Bayes Factors: * $BF_{10} > 3$ (moderate), ** $BF_{10} > 10$ (strong), and *** $BF_{10} > 100$ (extreme).} 
    \label{fig:prevalence_recognisability}
\end{figure}

So far, we have seen that when participants are shown MCQ options and asked to select the most recognisable, they show a tendency to select the option that is less prevalent in corpora. To further explore this, we take a closer look at the correct answers of the MCQs, specifically how often they are considered more recognisable than the distractors (Figure~\ref{fig:prevalence_recognisability}). In line with Table~\ref{tab:recognition_prevalence_corr}, we find that for most sets, the correct answer is not considered more recognisable than the distractors. Furthermore, for the Biopsychology questions, and especially the sets with LLM-generated distractors, we find that the recognisability of the correct answers decreases when they are more prevalent in corpora. These results point toward the possibility that LLMs might be generating distractors that are more prevalent in corpora, but less recognisable than human-generated distractors. The broader implications of these findings are further discussed in Section~\ref{sec:discussion}.

\subsection{Corpus Prevalence as a Proxy for Question Difficulty}
\label{sec:corpus_prevalene_question_difficulty_corr}
To explore whether the relative corpus prevalence of the correct answer compared to the distractors is indicative of MCQ difficulty, we use the difficulty measures (i.e., the proportion of students who answered each item correctly) available for the Biopsychology and Immunopharmacology question sets using the original distractors. As in Section \ref{sec:corpus_prevalence_and_recognisability}, we evaluate the correlation between corpus prevalence and difficulty using Kendall's Tau, applying a Bonferroni correction to adjust for multiple comparisons. Overall, corpus prevalence does not correlate with item difficulty. Although we observed a very weak, statistically significant correlation for in-context prevalence within Wikipedia (Table~\ref{tab:difficulty_correlation}), we conclude that the relative prevalence of the correct answer is not a sufficient predictor of item difficulty.

\begin{table}[h!]
\caption{Kendall's $\tau$ correlation between relative in- and out-of-context corpus prevalence of the correct answer and item difficulty. Significant results after Bonferroni correction for multiple comparisons are denoted with an asterisk~($\alpha = 0.013$).}
\centering
\setlength{\tabcolsep}{6pt}
\begin{tabular}{lllccc}
\toprule
\textbf{Dataset} & \textbf{Prevalence} & \textbf{Corpus} & \textbf{$N$} & \textbf{$\tau_b$} & \textbf{$p$-value}\\
\midrule
\multirow{4}{*}{\shortstack[l]{\textbf{Biopsychology} \\ (3 options)}} 
 & In-Context          & Wikipedia        & 319 & \phantom{-}0.1223 & $2.58 \times 10^{-3}(*)$  \\
 & In-Context          & Textbook & 319 & \phantom{-}0.0619 & $1.28 \times 10^{-1}$ \phantom{(*)}   \\
 & Out-Of-Context      & Wikipedia        & 319 & -0.0401           & $3.27 \times 10^{-1}$ \phantom{(*)}   \\
 & Out-Of-Context      & Textbook & 319 & \phantom{-}0.0114 & $7.80 \times 10^{-1}$  \phantom{(*)}  \\
\midrule
\multirow{4}{*}{\shortstack[l]{\textbf{Biopsychology} \\ (4 options)}} 
 & In-Context          & Wikipedia        & 278 & \phantom{-}0.1375 & $1.59 \times 10^{-3}(*)$\\
 & In-Context          & Textbook & 278 & \phantom{-}0.0616 & $1.58 \times 10^{-1}$ \phantom{(*)}  \\
 & Out-Of-Context      & Wikipedia        & 278 & \phantom{-}0.0254 & $5.66 \times 10^{-1}$ \phantom{(*)}   \\
 & Out-Of-Context      & Textbook & 278 & -0.0465           & $2.93 \times 10^{-1}$ \phantom{(*)}  \\
\midrule
\multirow{4}{*}{\shortstack[l]{\textbf{Immunopharmacology} \\ (4 options)}} 
 & In-Context          & Wikipedia        & 489 & \phantom{-}0.0863 & $8.95 \times 10^{-3}(*)$\\
 & In-Context          & Textbook   & 489 & -0.0037           & $9.13 \times 10^{-1}$ \phantom{(*)}\\
 & Out-Of-Context      & Wikipedia        & 489 & \phantom{-}0.0087 & $7.96 \times 10^{-1}$ \phantom{(*)} \\
 & Out-Of-Context      & Textbook   & 489 & \phantom{-}0.0187 & $5.77 \times 10^{-1}$ \phantom{(*)}    \\
\bottomrule
\end{tabular}
\label{tab:difficulty_correlation}
\end{table}

\section{Discussion}
\label{sec:discussion}
In this work we explored a computational method of quantifying the corpus prevalence of MCQ options leveraging semantic text embeddings. We found that the prevalence of the correct answer, relative to the distractors, is significantly higher, even when not accounting for the question (\textbf{RQ1}). Furthermore, this finding does not only hold for expert-created distractors, but can also be observed using LLM-generated distractors (\textbf{RQ2}). Lastly, we found that the relative corpus prevalence of MCQ options does not seem to correlate with how recognisable they are to the general public; in fact, we observed the opposite in some cases (\textbf{RQ3}).

A consistent finding across all experiments is the discrepancy in corpus prevalence depending on the retrieval corpus used. While prevalence differences are pronounced within the Wikipedia and Textbook corpora, they are negligible using BEIR. This suggests that the degree of specialisation and domain relevance of the corpus is much more important than its size alone. From an accessibility perspective, this is encouraging, as retrieving passages from small corpora is much more computationally efficient than doing so with large corpora. In the broader context, our findings with regards to the corpus choice highlights the care that needs to be given in selecting appropriate datasets when using computational approaches for education. 

Interestingly, corpus prevalence does not appear to correlate with higher relative recognisability. To understand why, we can look to the inherent nature of text corpora and Grice's maxim of quantity, which posits that utterances provide only as much information as necessary \citep{grice1975logic}. Consequently, we hypothesise that universally recognised concepts are explicitly mentioned less frequently than more specialised or novel terms. This aligns with the findings of Lin et al., who observed that \enquote{green banana} appears 332\% more often in the Google Books Ngram Corpus than \enquote{yellow banana} \citep{lin-etal-2012-syntactic}. More generally, and especially with the broader field of education in mind, this reinforces the idea that text corpora --- and by extension, the LLMs trained on them --- do not accurately simulate human behaviour.

Shifting the focus to MCQ evaluation, we found that the relative corpus prevalence of the correct answer compared to the distractors is only a weak predictor of difficulty (an aspect of MCQ quality). At the same time, and in connection to the $19$ \enquote{Item Writing Flaws}~\citep{Tarrant2006}, we argue that any difference in corpus prevalence between the options is undesirable, as MCQ options should not differ in this seemingly arbitrary manner. More broadly, this work raises questions about how MCQs are conceptualised in the first place, as our findings suggest that educators generate answers (and by extension, questions) that are somehow different from the distractors.

Lastly, reflecting on the fact that LLM capabilities are often evaluated through their performance on MCQs, the current work raises concerns regarding the true baseline performance on these evaluation datasets. This is in line with the findings of \citet{balepur-etal-2024-artifacts}, who found that LLMs, when given only the MCQ options without the question stem, can guess the correct answer with an accuracy significantly higher than random chance. More broadly, this is just one aspect where evaluating LLMs with MCQs falls short, alongside their sensitivity to arbitrary changes in the prompt, such as whether there is a space character at the end \citep{mind_the_gap} or the order of the options \citep{pezeshkpour2023large, wei2024unveiling}.

\section{Limitations}
\label{sec:limitations}
Because the experiments focused on factual knowledge, the datasets likely highlight differences in the corpus prevalence of specific MCQ options, specifically because these options are often technical terms. We expect that these findings would be more subtle in reasoning-heavy domains like mathematics. However, arguably, MCQs are inherently not the best way to assess reasoning skills.

Moreover, in Section~\ref{sec:behavioural_experiment}, we evaluated whether there is a relationship between corpus prevalence and recognisability. To evaluate recognisability, we recruited participants from Prolific, an online platform. While we filtered out data from participants who failed attention checks throughout the experiment, the remaining participants represent the general public rather than domain experts. Unfortunately, it was not feasible to conduct the experiment with students enrolled in these courses, but it would be interesting to see if they would respond differently. We believe this might be the case due to the different ways in which experts and non-experts relate to words. For example, for an MCQ with the options \enquote{cone}, \enquote{fovea}, and \enquote{choroid}, \enquote{cone} is recognisable to non-experts simply because it is used in broader contexts (e.g., traffic cone).

Finally, our methodology used a simple instruction prompt to generate distractors using LLMs (Figure~\ref{fig:distractor_generation_prompt}). This contrasts with extensive prior work regarding strategies for creating pedagogically sound distractors~\citep{alhazmi-distractor-survey}. Nevertheless, given the time constraints faced by educators, we argue that a simple prompt is more representative of real-world usage than sophisticated distractor-generation pipelines.

\appendix

\section{Relative In-Context Corpus Prevalence}
\label{sec:in_context_prevalence}
As described in Section \ref{sec:introduction}, we draw a distinction between in- and out-of-context corpus prevalence. Using our methodology, we can capture in- and out-of-context prevalence by modifying the passage retrieval query, to either include or exclude the question stem. As expected, in-context corpus prevalence favours the correct answer even more compared to the results seen in Section~\ref{sec:relative_corpus_prevalence}, using the Wikipedia and Textbook Corpora. Using BEIR this pattern is not consistently observed.

\begin{figure}[h!]
    \centering
    \includegraphics[width=1\linewidth]{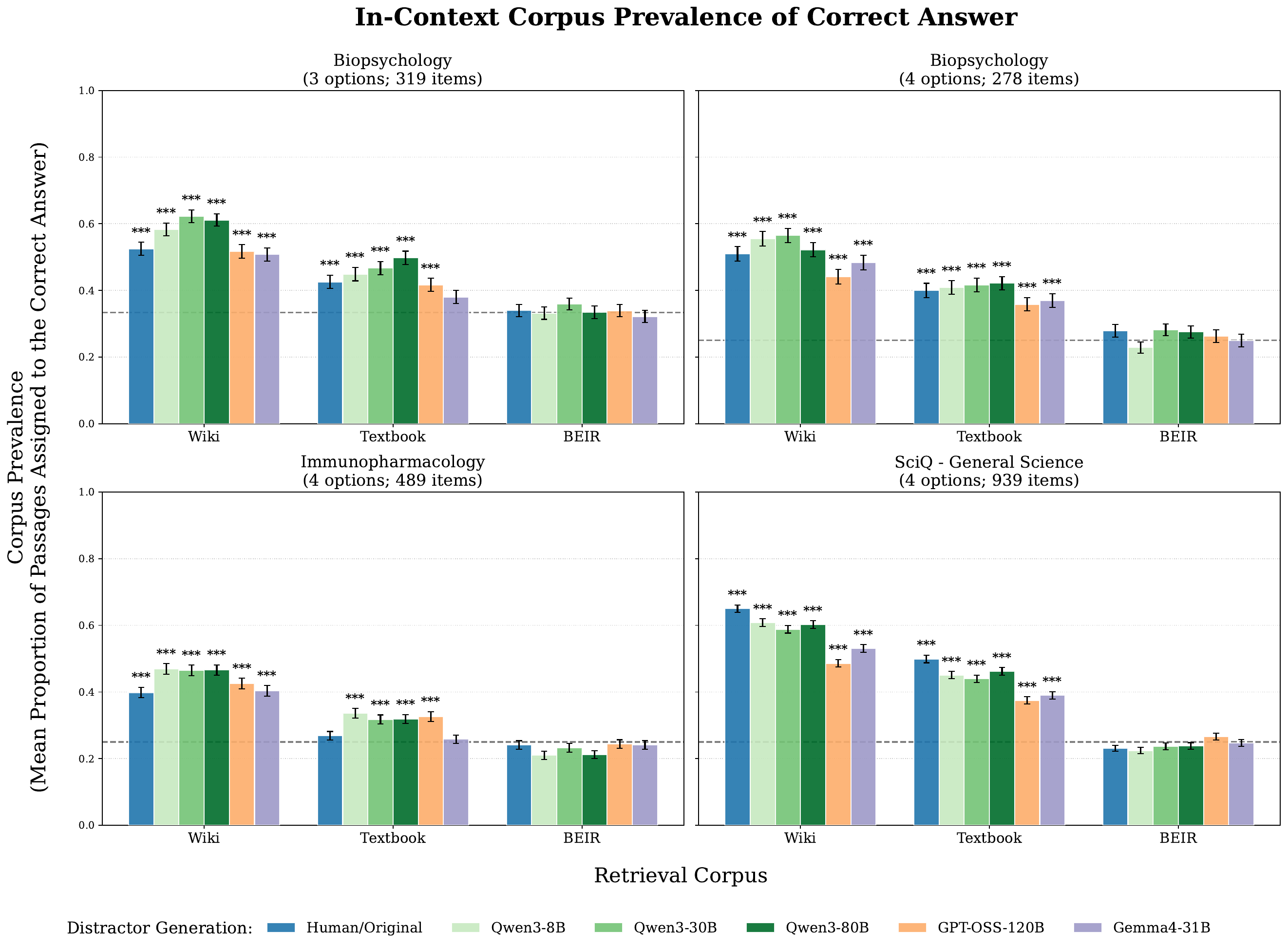}
    \caption{Relative in-context corpus prevalence of the correct answer of MCQs varying the distractor generation method and the retrieval corpus. Dashed lines indicate the baseline random-assignment proportion. Asterisks indicate the strength of evidence for the alternative hypothesis based on Bayes Factors: * $BF_{10} > 3$ (moderate), ** $BF_{10} > 10$ (strong), and *** $BF_{10} > 100$ (extreme).}
    \label{fig:corpus_prevalence_in_context}
\end{figure}

\begin{table}[h!]
\caption{Proportion of items where the correct answer has higher in-context prevalence in the Wikipedia than the distractors. The increase (or decrease) over the baseline (33.3\% for 3 options, 25.0\% for 4 options) is denoted in brackets.}
\centering
\setlength{\tabcolsep}{5pt}
\begin{tabular}{ccccc}
\toprule
\multirow{2}{*}{\textbf{Distractors Generator}} & \multicolumn{4}{c}{\textbf{\% Correct Answer as Most Prevalent (Wikipedia)} $\uparrow$} \\
\cmidrule(lr){2-5}
 & \shortstack[c]{\textbf{Biopsychology} \\ (3 options)} & \shortstack[c]{\textbf{Biopsychology} \\ (4 options)} & \shortstack[c]{\textbf{Immunopharmacology} \\ (4 options)} & \shortstack[c]{\textbf{SciQ} \\ (4 options)} \\
\midrule
Human/Original & 53.6\% {\scriptsize (+20.3\%)} & 48.2\% {\scriptsize (+23.2\%)} & 38.2\% {\scriptsize (+13.2\%)} & 68.7\% {\scriptsize (+43.7\%)} \\
Gemma4-31b     & 52.0\% {\scriptsize (+18.7\%)} & 50.0\% {\scriptsize (+25.0\%)} & 39.5\% {\scriptsize (+14.5\%)} & 53.4\% {\scriptsize (+28.4\%)} \\
GPT-120b       & 51.4\% {\scriptsize (+18.1\%)} & 42.4\% {\scriptsize (+17.4\%)} & 43.1\% {\scriptsize (+18.1\%)} & 48.3\% {\scriptsize (+23.3\%)} \\
Qwen3-8b       & 59.2\% {\scriptsize (+25.9\%)} & 57.6\% {\scriptsize (+32.6\%)} & 48.3\% {\scriptsize (+23.3\%)} & 63.5\% {\scriptsize (+38.5\%)} \\
Qwen3-30b      & 64.3\% {\scriptsize (+31.0\%)} & 59.4\% {\scriptsize (+34.4\%)} & 46.4\% {\scriptsize (+21.4\%)} & 60.8\% {\scriptsize (+35.8\%)} \\
Qwen3-80b      & 64.6\% {\scriptsize (+31.3\%)} & 54.7\% {\scriptsize (+29.7\%)} & 47.4\% {\scriptsize (+22.4\%)} & 62.9\% {\scriptsize (+37.9\%)} \\
\bottomrule
\end{tabular}
\label{tab:accuracy_majority_in_context}
\end{table}

\section{Varying the Number of Retrieved Passages}
\label{sec:number_of_retrieved_passages}
\begin{figure}[h!]
    \centering
    \includegraphics[width=0.75\linewidth]{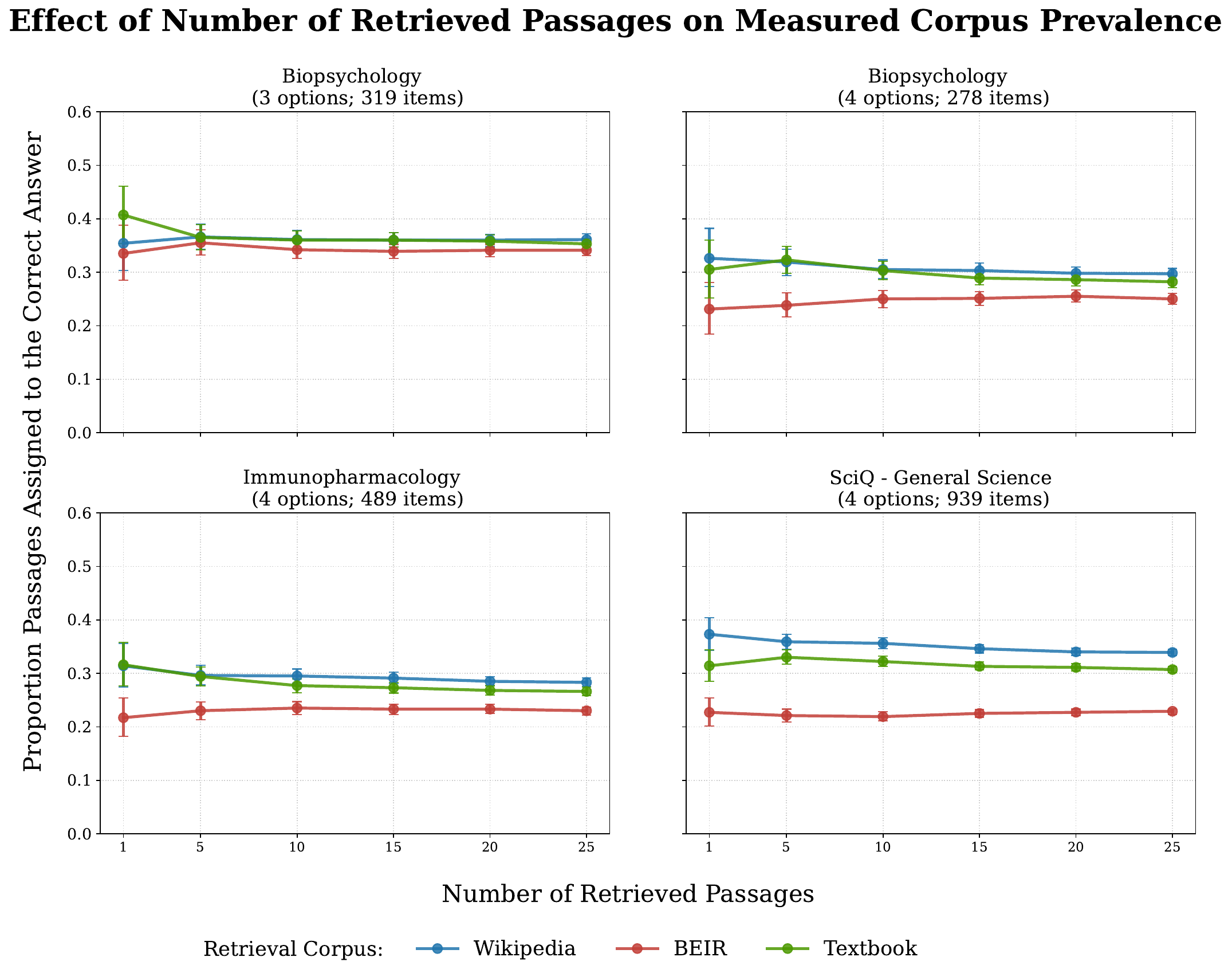}
    \caption{Impact of the number of retrieved passages on measured corpus prevalence. The presented data reflects out-of-context prevalence using the original, human-generated distractors.}
    \label{fig:num_of_docs}
\end{figure}

In this analysis, we explore the effect of the number of retrieved passages from the corpora on the measured prevalence. This serves two purposes: to establish an appropriate passage count for all experiments, and to assess whether our methodology is affected by topic specialization. Presumably, retrieving a large number of passages for a highly specialized query might return irrelevant passages, thereby skewing our measurement. To test this, we measured the relative out-of-context prevalence of the correct answer across all datasets using the original human-created distractors (Figure~\ref{fig:num_of_docs}). We found that retrieving more than five passages does not meaningfully alter the prevalence measure, but significantly increases computational cost. This finding holds across both niche subjects (e.g., Immunopharmacology) and general subjects (e.g., SciQ).

 \bibliographystyle{elsarticle-harv} 
 \bibliography{bibliography}

\end{document}